\documentclass{article}

\usepackage{PRIMEarxiv}

\usepackage[utf8]{inputenc} 
\usepackage[T1]{fontenc}    
\usepackage{hyperref}       
\usepackage{url}            
\usepackage{booktabs}       
\usepackage{amsfonts}       
\usepackage{nicefrac}       
\usepackage{microtype}      
\usepackage{lipsum}
\usepackage{fancyhdr}       
\usepackage{graphicx}       
\graphicspath{{media/}}     
\usepackage{natbib}
\usepackage{ragged2e}
\usepackage{tabularx}
\usepackage{caption}
\usepackage[section]{placeins}
\newcolumntype{Y}{>{\RaggedRight\arraybackslash}X}
\usepackage{float}
\usepackage{amsmath}
\usepackage{amssymb}
\title{Purrturbed but Stable: Human-Cat Invariant Representations Across CNNs, ViTs and Self-Supervised ViTs
}

\author{
  Arya Shah \\
  Department of Computer Science, \\
  Indian Institute of Technology\\
Gandhinagar, India \\
  \texttt{arya.shah@iitgn.ac.in} \\
   \And
  Vaibhav Tripathi \\
  Department of Cognitive and Brain Sciences \\
  Indian Institute of Technology\\
Gandhinagar, India \\
  \texttt{vaibhav.tripathi@iitgn.ac.in} \\
}

\begin{document}
\maketitle
\begin{abstract}
Cats and humans differ in ocular anatomy. Most notably, Felis Catus (domestic cats) have vertically elongated pupils linked to ambush predation; yet, how such specializations manifest in downstream visual representations remains incompletely understood. We present a unified, frozen-encoder benchmark that quantifies feline-human cross-species representational alignment in the wild, across convolutional networks, supervised Vision Transformers, windowed transformers, and self-supervised ViTs (DINO), using layer-wise Centered Kernel Alignment (linear and RBF) and Representational Similarity Analysis, with additional distributional and stability tests reported in the paper. Across models, DINO ViT-B/16 attains the most substantial alignment (mean CKA-RBF $\approx0.814$, mean CKA-linear $\approx0.745$, mean RSA $\approx0.698$), peaking at early blocks, indicating that token-level self-supervision induces early-stage features that bridge species-specific statistics. Supervised ViTs are competitive on CKA yet show weaker geometric correspondence than DINO (e.g., ViT-B/16 RSA $\approx0.53$ at block8; ViT-L/16 $\approx0.47$ at block14), revealing depth-dependent divergences between similarity and representational geometry. CNNs remain strong baselines but below plain ViTs on alignment, and windowed transformers underperform plain ViTs, implicating architectural inductive biases in cross-species alignment. Results indicate that self-supervision coupled with ViT inductive biases yields representational geometries that more closely align feline and human visual systems than widely used CNNs and windowed Transformers, providing testable neuroscientific hypotheses about where and how cross-species visual computations converge.
We release our code and dataset on \href{https://github.com/aryashah2k/Purrturbed-but-Stable}{Github} for reference and reproducibility.
\end{abstract}
\section{Introduction}
\begin{figure*}[t]
  \centering
  \includegraphics[width=\textwidth]{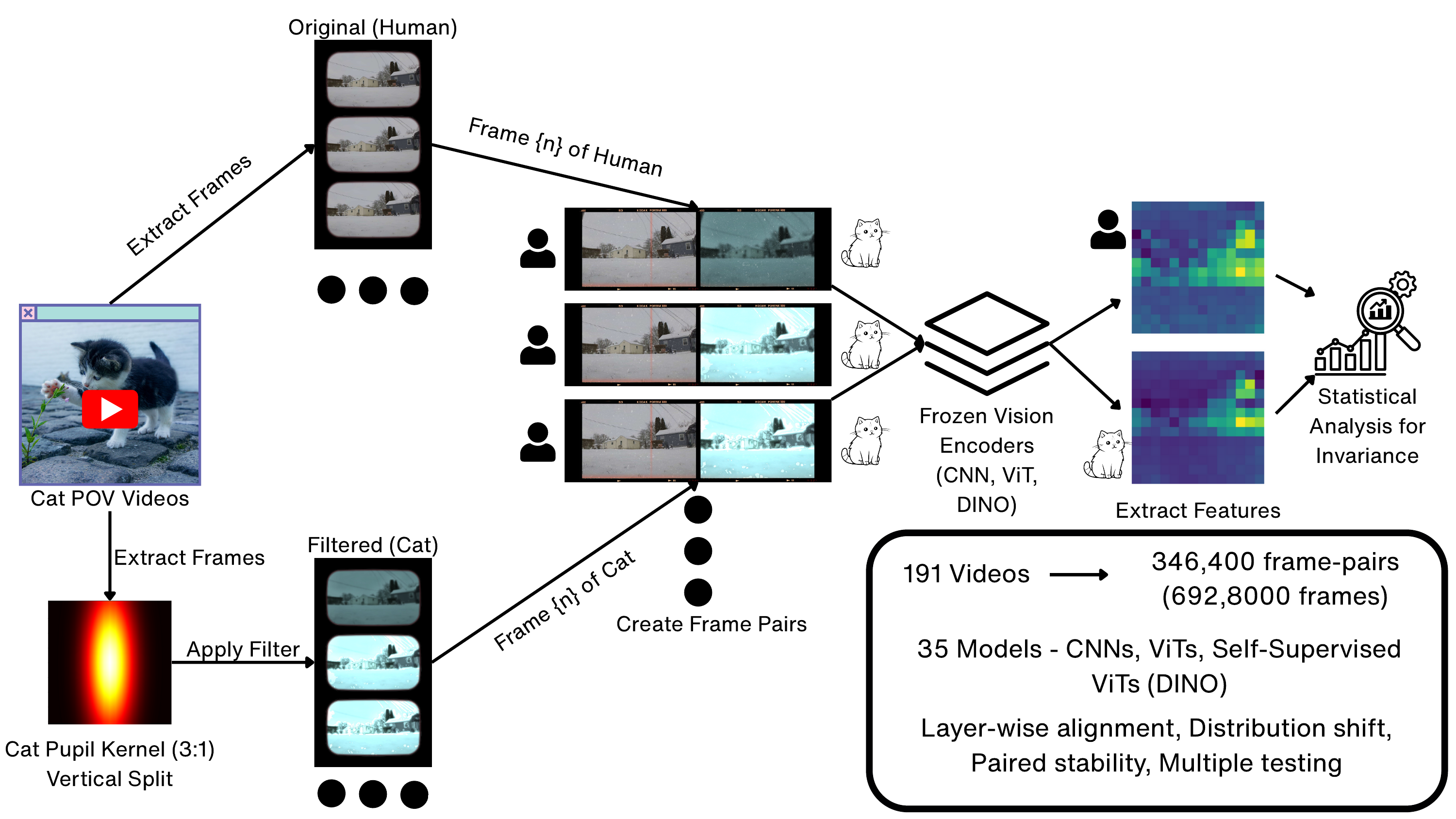}
  \caption{Process Overview for Measuring Human-Cat Invariant Representations in CNNs, ViTs and Self-Supervised ViTs. A total of 191 videos containing cat POV videos are sourced from the internet. Our biologically informed cat vision filter is applied to individual frames and we create pairs of original (human) vs. cat vision filtered frames which pass through a suite of frozen vision encoders and the extracted features are then subjected to statistical tests.}
  \label{fig:overview}
\end{figure*}
Representation learning has reshaped computer vision, with convolutional networks (CNNs) and Vision Transformers (ViTs) delivering strong features for transfer and understanding \citep{he2015deepresiduallearningimage,huang2018denselyconnectedconvolutionalnetworks,tan2020efficientnetrethinkingmodelscaling,liu2022convnet2020s,dosovitskiy2021imageworth16x16words,liu2021swintransformerhierarchicalvision}. A growing body of work evaluates learned representations by their geometry and cross-domain stability, using tools such as Centered Kernel Alignment (CKA) and Representational Similarity Analysis (RSA) to compare models and brains \citep{kornblith2019similarityneuralnetworkrepresentations,10.3389/neuro.06.004.2008}. However, despite interest in biological plausibility and cross-species comparisons (e.g., alignment studies between artificial and biological vision \citep{yamins2014performance, schrimpf2021brain,Schrimpf2018BrainScoreWA}), there is limited, systematic evidence on how modern encoders align visual representations across species differences that arise from distinct ocular and ecological constraints.

This work addresses that gap with a unified, frozen-encoder benchmark that quantifies cross-species representational alignment between cats and humans. We compare paired feline-human images across widely used CNN families (ResNet, DenseNet, EfficientNet, ConvNeXt, MobileNet) \citep{howard2017mobilenetsefficientconvolutionalneural, sandler2019mobilenetv2invertedresidualslinear, howard2019searchingmobilenetv3}, supervised ViTs (ViT-B/L), windowed transformers (Swin-T/S/B), and self-supervised ViTs (DINO and DINOv2/v3) \citep{caron2021emergingpropertiesselfsupervisedvision,oquab2024dinov2learningrobustvisual, siméoni2025dinov3}. To ensure comparability and statistical rigor, we measure alignment layer-wise using CKA (linear and RBF) and RSA/Mantel, and we probe distributional differences via MMD, Energy distance, and 1‑Wasserstein, alongside paired stability tests with Benjamini-Hochberg FDR control.

At a high level, the approach is simple: freeze encoders, extract per-layer features on paired feline-human images, vectorize features consistently (tokens for ViTs, global average pooling for CNN-like maps), and compute alignment and shift statistics per model and layer. This design isolates representational similarities attributable to the trained feature spaces themselves, rather than to downstream heads or fine-tuning protocols; by keeping models frozen, we test inherent alignment capacity without confounding effects of task-specific fine-tuning or domain adaptation.

On our dataset, self-supervised ViTs exhibit the strongest cross-species alignment. In particular, DINO ViT‑B/16 attains mean CKA‑RBF 0.814, mean CKA‑linear 0.745, and mean RSA (Mantel) 0.698 with peaks at early blocks; supervised ViTs are competitive on CKA‑linear but show weaker geometric correspondence (e.g., ViT‑B/16 RSA $\approx0.53$; ViT‑L/16 $\approx0.47$). CNNs remain strong baselines yet generally trail plain ViTs on alignment (e.g., ResNet‑50 mean CKA‑linear 0.663, RSA 0.488), and windowed transformers (Swin) underperform plain ViTs on average. DINOv2/v3 variants show high early-layer maxima but lower means than DINO ViT‑B/16. These findings suggest that token-level self-supervision plus ViT inductive biases yield early-stage features that better bridge species-specific statistics than widely used CNNs and windowed transformers.

In this work, we contribute (1) A unified, frozen-encoder benchmark and analysis pipeline for cross-species representational alignment spanning CNNs (ResNet, DenseNet, EfficientNet, ConvNeXt, MobileNet), supervised ViTs (ViT-B/L), windowed transformers (Swin-T/S/B), and self-supervised ViTs (DINO, DINOv2/v3). (2) A layer-wise, model-agnostic evaluation using CKA (linear/RBF) and RSA/Mantel with distribution-shift and paired-stability tests under BH-FDR, enabling precise comparisons across heterogeneous architectures. (3) Quantitative evidence that self-supervised ViTs, especially DINO ViT‑B/16, achieve the strongest feline-human alignment with early-layer peaks, while supervised ViTs, CNNs, and Swin lag to varying degrees. (4) A new in-the-wild dataset of 191 videos yielding $\approx300,000$ paired human-cat frames plus a \emph{biologically informed cat-vision filter} (novel, first-of-its-kind implementation), enabling controlled cross-species benchmarking; we document curation and filtering details.

\section{Related Work}
\par Prior work spans four themes relevant to our study: (i) representation similarity and analysis methods, (ii) architectural families for vision and self-supervision, (iii) alignment of artificial and biological vision, and (iv) benchmarks probing invariances and frozen-encoder evaluations. We review each and situate our contribution: a unified, \emph{frozen-encoder}, per-layer cross-species analysis across CNNs, supervised ViTs, and DINO-style ViTs with comprehensive statistics.

\subsection{Representation Similarity and Analysis Methods}
\par RSA compares internal representational geometries via RDMs across models, brains, and behavior \citep{10.3389/neuro.06.004.2008}. Subspace methods align activations: SVCCA (PCA + CCA) \citep{raghu2017svccasingularvectorcanonical} and PWCCA (projection-weighted) \citep{morcos2018insights}. CKA offers invariant, empirically stable similarity across architectures and runs \citep{kornblith2019similarityneuralnetworkrepresentations}. Orthogonal Procrustes supports rotation-constrained alignment for complementary perspectives \citep{Schnemann1966AGS}. To capture distributional differences beyond similarity, kernel two-sample tests (MMD) \citep{JMLR:v13:gretton12a}, energy distance \citep{Rizzo2016EnergyD}, and Wasserstein distances quantify shifts between domains or species.
\par Reliability and inference are integral to RSA: the Mantel test assesses significance via matrix permutations \citep{mantel1967detection}; noise ceilings and reliability checks contextualize absolute scores \citep{WALTHER2016188}. For CKA, its HSIC foundation links representational similarity to RKHS dependence, guiding linear vs. RBF choices by expected nonlinearities \citep{gretton2008kernelmethodtwosampleproblem}.
\par We use \emph{frozen-encoder, per-layer} alignment with CKA (linear/RBF) and RSA/Mantel, plus MMD, Energy, and 1‑Wasserstein, employing paired tests under BH-FDR to jointly quantify similarity and shift across species.

\subsection{ConvNets, Vision Transformers, and Self-Supervised ViTs}
\par CNN encoders include residual \citep{he2015deepresiduallearningimage}, dense \citep{huang2018denselyconnectedconvolutionalnetworks}, compound-scaled \citep{tan2020efficientnetrethinkingmodelscaling}, modernized \citep{liu2022convnet2020s}, and mobile-efficient designs \citep{howard2017mobilenetsefficientconvolutionalneural, sandler2019mobilenetv2invertedresidualslinear, howard2019searchingmobilenetv3}. ViT introduces tokenized global attention \citep{dosovitskiy2021imageworth16x16words}; Swin adds hierarchical shifted windows \citep{liu2021swintransformerhierarchicalvision}. Self-supervised ViTs (DINO, DINOv2/v3) yield strong token-level features without labels and scale effectively \citep{caron2021emergingpropertiesselfsupervisedvision, oquab2024dinov2learningrobustvisual, siméoni2025dinov3}. Frozen-encoder transfer is competitive, yet \emph{cross-species}, per-layer alignment remains underexplored; our evaluation compares CNNs, supervised ViTs, and DINO-style ViTs in a unified frozen setting.
\par Data-efficient supervised ViTs (DeiT) leverage distillation/augmentation \citep{touvron2021deit}; masked image modeling (BEiT) improves token-level pretexts \citep{bao2021beit}. Contrastive and non-contrastive SSL (MoCo v2, BYOL) establish strong frozen-transfer baselines \citep{chen2020improvedbaselinesmomentumcontrastive,grill2020bootstraplatentnewapproach}, and token-centric iBOT extends masked prediction with instance/patch tasks \citep{zhou2022ibot}. These trends position self-supervised ViTs as promising for invariant representations, aligning with our findings.

\subsection{Aligning Artificial and Biological Vision}
\par Performance-optimized hierarchical models predict high-level ventral stream responses \citep{yamins2014performance}; Brain-Score aggregates neural and behavioral benchmarks \citep{Schrimpf2018BrainScoreWA,schrimpf2021brain}. RSA unifies representational geometry comparisons across species and modalities \citep{10.3389/neuro.06.004.2008}, and CKA offers desirable invariances for encoder comparisons \citep{kornblith2019similarityneuralnetworkrepresentations}. Time-resolved RSA with MEG/EEG links model layers to temporal dynamics \citep{Cichy2014}.
\par Goal-driven modeling ties task optimization to cortical predictivity \citep{Yamins2016}, extending to early visual areas under naturalistic stimuli \citep{Cadena2019}, supporting learned encoders as testable hypotheses about neural coding.
\par Ecological factors shape species vision; vertical slit pupils in ambush predators reflect niche-specific constraints \citep{Banks2015}, motivating cross-species comparisons on in-the-wild data and biologically informed prefilters. We use a cat-vision filter to approximate species-specific inputs and assess frozen, layer-wise alignment.
\par Unlike brain-predictivity or within-species behavior studies, we measure \emph{cross-species representational alignment} directly on feline-human imagery, combining per-layer RSA/CKA with distributional tests.

\subsection{Benchmarks, Invariances, and Frozen-Encoder Evaluations}
\par Benchmarks expose invariance and robustness gaps: ImageNet-trained CNNs often favor texture over shape \citep{geirhos2022imagenettrainedcnnsbiasedtexture}; Stylized-ImageNet encourages shape bias \citep{geirhos2022imagenettrainedcnnsbiasedtexture}. Corruption benchmarks (CIFAR-10-C, ImageNet-C) quantify robustness \citep{hendrycks2019benchmarkingneuralnetworkrobustness}. ImageNet-A/O/R probe natural adversarial, OOD, and rendition shifts \citep{hendrycks2021naturaladversarialexamples}. Broader efforts measure real-world distribution shift \citep{taori2020measuring,koh2021wildsbenchmarkinthewilddistribution}. SSL shows strong frozen transfer (SimCLR, BYOL, MAE) \citep{chen2020simclr,grill2020bootstraplatentnewapproach,he2022mae}.
\par We complement this literature by targeting \emph{cross-species} invariance with a unified frozen-encoder, layer-wise pipeline across CNNs, supervised ViTs, and DINO/DINOv2/v3, quantifying RSA/CKA and MMD, Energy, and 1‑Wasserstein with paired stability tests and BH-FDR; our in-the-wild feline-human dataset and cat-vision filter add ecological validity.
\par Distinct from within-domain/model comparisons or single-metric studies, we analyze paired cross-species data across heterogeneous frozen encoders with RSA/CKA and distributional tests under FDR control, enabled by our dataset and biologically informed filter.

\section{Methodology}
We study cross-species representational invariance by comparing layer-wise activations from diverse vision encoders on paired views of the same scenes captured in human and feline domains. All encoders are kept frozen to isolate the inductive biases of the pretrained features, enabling apples-to-apples comparisons across architectures and training paradigms.

Our methodology comprises six stages: (1) constructing a strictly paired dataset with filename-level alignment and integrity checks; (2) standardizing inputs using each encoder's canonical preprocessing; (3) extracting intermediate activations at semantically comparable layers along the processing hierarchy; (4) vectorizing activations to a common representation to remove architectural idiosyncrasies; (5) quantifying alignment and shift with complementary metrics such as Centered Kernel Alignment (CKA) \citep{kornblith2019similarityneuralnetworkrepresentations}, Representational Similarity Analysis (RSA) and Mantel permutation testing \citep{10.3389/neuro.06.004.2008,mantel1967detection}, plus Maximum Mean Discrepancy (MMD) \citep{JMLR:v13:gretton12a}, Energy distance \citep{Rizzo2016EnergyD}, and a projected 1-Wasserstein distance under false-discovery-rate control; and (6) reporting layer-wise summaries and diagnostic visualizations to contextualize effects across families. Design choices reflect established best practices in representation analysis: frozen-encoder evaluation to avoid confounds from fine-tuning; layer-wise probes to localize where alignment emerges; vectorization via global average pooling for map features and class-token or token-mean for tokenized features \citep{dosovitskiy2021imageworth16x16words,liu2021swintransformerhierarchicalvision}; and complementary geometric and distributional tests to capture both similarity and shift.

\subsection{Dataset}
We curate a paired image dataset to enable controlled cross-species comparisons on the same scenes. POV(point-of-view) videos of domestic cats with a camera strapped to their neck are collected from the internet and temporally aligned and decomposed into frames; images are then paired at the filename level to ensure one-to-one correspondences. Pairs with missing counterparts, corrupted files, or non-RGB formats are excluded. Each pair is assigned a stable identifier for reproducibility and joined consistently across outputs. Our curated dataset turns out to be 191 videos large, consisting of over 300,000 frame-pairs. To minimize confounds, we adopt three practices: (i) mirroring directory structures across domains so that every human frame has at most one feline counterpart; (ii) enforcing identical preprocessing pipelines per model family (resolution, normalization) to avoid input-induced artifacts in representational geometry; and (iii) evaluating only exact pairs for which both domains are present. This conservative pairing mitigates label leakage or scene-mismatch issues common in out-of-distribution evaluations \citep{taori2020measuring,koh2021wildsbenchmarkinthewilddistribution}.

We report all results with per-layer sample counts to make coverage explicit and avoid over-interpreting layers with few matched pairs. No personal identifiers are present, and images are drawn from public, in-the-wild recordings with standard research use; we analyze representational statistics and do not perform identity recognition. For each encoder, images are resized and normalized according to its canonical evaluation recipe (e.g., ImageNet-style center crop for convolutional networks; patch embedding resolutions for transformer encoders). Analyses are performed on the full paired corpus; for ablations we optionally downsample with a fixed random seed to a ``golden'' subset to test stability of conclusions. We avoid fine-tuning to isolate encoder invariances. Paired cross-domain datasets enable stronger causal interpretations of representational alignment than independent draws, as they hold scene content fixed while varying the domain. This design complements robustness benchmarks that evaluate distribution shifts without strict pairing \citep{hendrycks2019benchmarkingneuralnetworkrobustness,taori2020measuring,koh2021wildsbenchmarkinthewilddistribution} and helps localize where along the processing hierarchy cross-species invariances are preserved.

We situate our corpus relative to existing resources involving animal or egocentric visual data. To our knowledge, public datasets rarely provide strictly paired, same-scene human-animal imagery; most focus on single-species egocentric capture or unpaired wildlife monitoring. Table~\ref{tab:related_datasets} summarizes representative datasets and highlights how our paired, cross-species framing differs.

\begin{table}[t]
\centering
\caption{Representative related datasets. Unlike these resources, our dataset provides strictly paired, same-scene images across human and feline domains to enable per-layer cross-species alignment analyses.}
\label{tab:related_datasets}
\setlength{\tabcolsep}{4pt}
\renewcommand{\arraystretch}{1.1}
\begin{tabularx}{\linewidth}{@{}%
  >{\RaggedRight\arraybackslash}p{2.9cm}%
  >{\RaggedRight\arraybackslash}p{1.6cm}%
  >{\RaggedRight\arraybackslash}p{2.1cm}%
  >{\RaggedRight\arraybackslash}p{2.1cm}%
  >{\RaggedRight\arraybackslash}X@{}}
\toprule
Dataset (year/venue) & Species & Modality & Paired same-scene? & Key difference vs. ours \\
\midrule
CatCam \citep{Betsch2004} & Cat & Egocentric video & No & Natural egocentric feline videos; no human counterpart; used for natural video statistics. \\
DogCentric Activity \citep{yumi2014first} & Dog & Egocentric video & No & First-person animal activity recognition; single species, unpaired with humans. \\
EgoPet \citep{bar2024egopetegomotioninteractiondata} & Pet animals & Egocentric video & No & Animal egocentric interactions/locomotion benchmarks; not cross-species paired. \\
EGO4D \citep{grauman2022ego4dworld3000hours} & Human & Egocentric video & No & Large-scale human egocentric dataset; provides human-only perspective and tasks. \\
WILDS iWildCam \citep{koh2021wildsbenchmarkinthewilddistribution} & Wildlife & Camera traps & No & Distribution shifts across camera locations and time; not paired same-scene, different sensing modality. \\
\bottomrule
\end{tabularx}
\end{table}

\subsection{Biologically Informed Cat-Vision Filter}
\captionsetup{font=small,labelfont=bf}

\begin{figure*}[t]
  \centering
  \includegraphics[width=\textwidth]{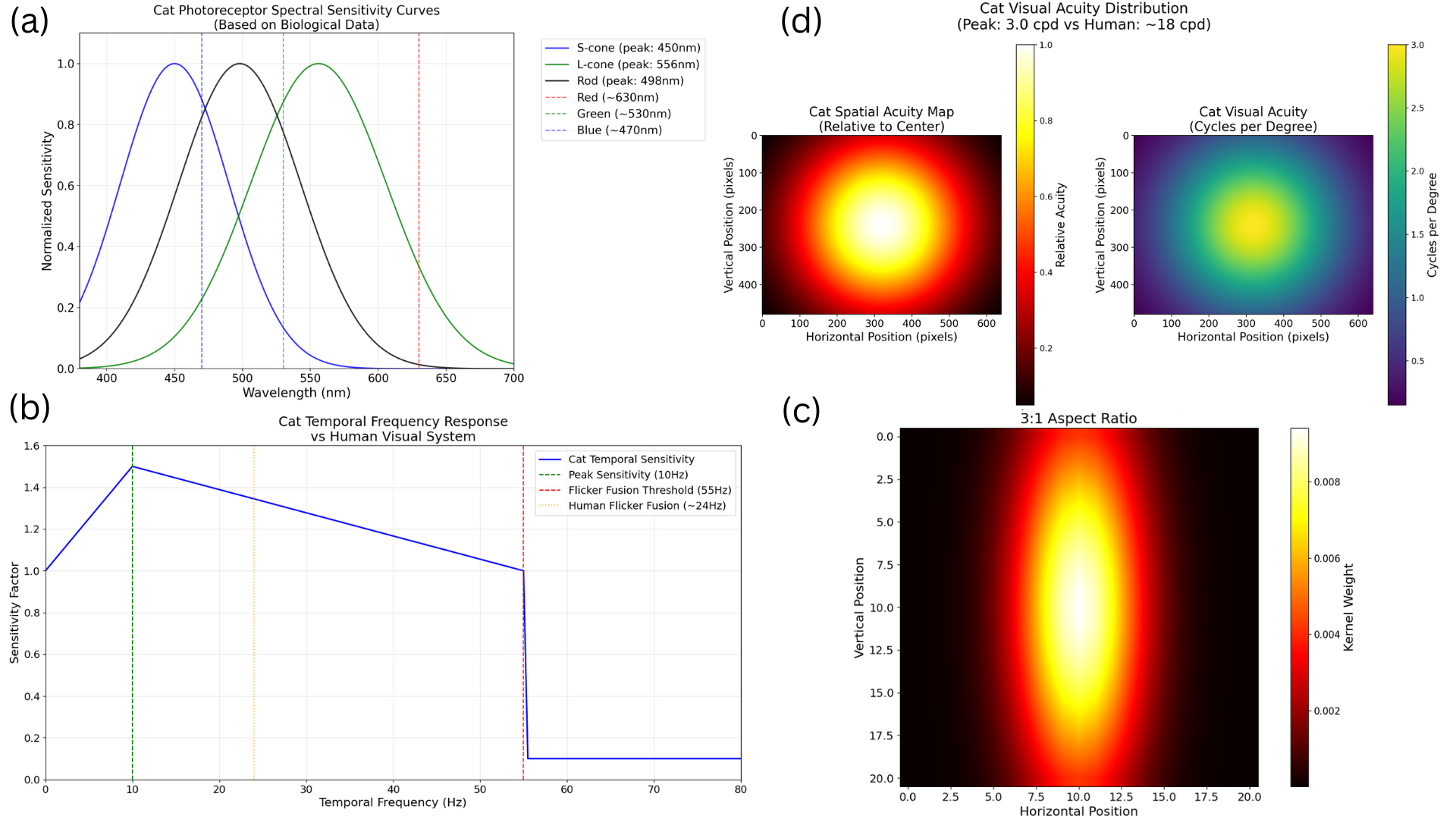}
  \caption{Moving counter-clockwise, Panel (a) depicts the cat's photoreceptor spectral sensitivity curves based on biological data, Panel (b) depicts the temporal frequency response of cat vs. human visual system, Panel (c) shows the cat's vertical slit pupil kernel in 3:1 aspect ratio. (d) shows the cat's spectral and visual acuity map based on our biologically informed implementation of cat's pupil, and Panel}
  \label{fig:catvision_filter}
\end{figure*}
To probe whether architectural invariances persist under species-specific optics and early vision, we apply a biologically informed image transformation that approximates key feline visual characteristics prior to feature extraction. The transformation models (i) spectral sensitivity with rod dominance and reduced long-wavelength sensitivity; (ii) spatial acuity and peripheral falloff; (iii) extended field-of-view distortions; (iv) temporal sensitivity and elevated flicker fusion; (v) motion sensitivity with horizontal bias; (vi) vertical-slit pupil optics; and (vii) a tapetum lucidum low-light enhancement. We interleave each component with its mathematical definition and refer back to equations where relevant.
Spectral sensitivity is modeled by smooth sensitivity curves for short- and long-wavelength cones and rods, with peaks near 450 nm, 550-556 nm, and \(\approx\)498-501 nm, respectively, consistent with electrophysiology of feline retina and ganglion cells \citep{Andrews1970,Loop1987,Guenther1993}. We define Gaussian-like sensitivity functions \(S(\lambda), L(\lambda), R(\lambda)\) (\eqref{eq:spectral_curves1},\eqref{eq:spectral_curves2},\eqref{eq:spectral_curves3}) and compute cone/rod activations per pixel, then combine them with rod-dominant weights (\eqref{eq:spectral_blend}). This blending reflects nocturnal/crepuscular behavior and attenuated red sensitivity \citep{zrenner1979blue}.
\begin{align}
 S(\lambda)&=\exp\!\Big(-\tfrac{(\lambda-\mu_S)^2}{2\sigma_S^2}\Big),\quad \mu_S\approx 450\,\text{nm}\label{eq:spectral_curves1},\\
 L(\lambda)&=\exp\!\Big(-\tfrac{(\lambda-\mu_L)^2}{2\sigma_L^2}\Big),\quad \mu_L\approx 556\,\text{nm} \label{eq:spectral_curves2},\\
 R(\lambda)&=\exp\!\Big(-\tfrac{(\lambda-\mu_R)^2}{2\sigma_R^2}\Big),\quad \mu_R\approx 498{-}501\,\text{nm}.\label{eq:spectral_curves3}
\end{align}
With nominal RGB wavelengths \((\lambda_R,\lambda_G,\lambda_B)\), per-pixel activations are
\begin{align}
 C_S(x)&=w_B\,S(\lambda_B)\,B(x)+w_G\,S(\lambda_G)\,G(x)+w_R\,S(\lambda_R)\,R(x),\\
 C_L(x)&=w_B\,L(\lambda_B)\,B(x)+w_G\,L(\lambda_G)\,G(x)+w_R\,L(\lambda_R)\,R(x),\\
 C_R(x)&=w_B\,R(\lambda_B)\,B(x)+w_G\,R(\lambda_G)\,G(x)+w_R\,R(\lambda_R)\,R(x).\label{eq:spectral_channels}
\end{align}
Rod dominance is enforced by
\begin{equation}
 I_{spec}(x)=\beta_S C_S(x)+\beta_L C_L(x)+\beta_R C_R(x),\quad \beta_R: \beta_S+\beta_L \approx 25:1,\label{eq:spectral_blend}
\end{equation}
and the tapetum lucidum \citep{Coles1971} is approximated downstream via a luminance-dependent gain with a blue-green tint (Eq.\ \eqref{eq:tapetum}).

Spatial acuity is reduced with a frequency-domain low-pass filter to approximate lower feline grating acuity. Specifically, we use a Gaussian transfer function \(H(u,v)\) (Eq.\ \eqref{eq:lowpass}) and reconstruct the filtered image by inverse Fourier transform, choosing \(\sigma_{lp}\) to achieve \(\approx\)1/6 of human high-contrast acuity.
\begin{equation}
 H(u,v)=\exp\!\Big(-\tfrac{u^2+v^2}{2\sigma_{lp}^2}\Big),\qquad I_{sp}(x)=\mathcal{F}^{-1}\{ H\cdot \mathcal{F}\{I_{spec}\}\}(x).\label{eq:lowpass}
\end{equation}

Geometric optics are approximated by a vertical-slit pupil and a broadened field of view. Vertical slits are linked to ambush predation and depth-of-field control \citep{Banks2015}. We include a mild barrel distortion and a center-surround acuity mask (Eqs.\ \eqref{eq:barrel}-\eqref{eq:acuitymask}) to emulate wider effective field and peripheral blur.
\begin{equation}
 r' = r\,(1+k_1 r^2 + k_2 r^4),\quad \tilde{x}' = \tfrac{r'}{\max(r,\,\epsilon)}\,\tilde{x},\quad x'=\Pi(\tilde{x}'),\label{eq:barrel}
\end{equation}
\begin{equation}
 A(r)=\frac{1}{1+\exp\big(\gamma\,(r-r_0)\big)},\qquad I_{fov}(x)=A(r(x))\,I_{sp}(x').\label{eq:acuitymask}
\end{equation}

Temporal processing emphasizes motion sensitivity in the \(\sim\)10 Hz band, with reduced gain beyond \(\approx\)50-60 Hz. We apply a temporal band-pass gain \(G(f)\) in the Fourier domain (Eq.\ \eqref{eq:temporal}), with peak \(f_0\approx 10\,\text{Hz}\) and flicker-fusion cutoff \(f_{ff}\approx 55\,\text{Hz}\).
\begin{equation}
 I_{tmp}(t)=\mathcal{F}_t^{-1}\big\{G(f)\,\widehat{I}(f)\big\}(t),\qquad G(f)=\exp\!\Big(-\tfrac{(f-f_0)^2}{2\sigma_f^2}\Big)\,\mathbf{1}[f\le f_{ff}].\label{eq:temporal}
\end{equation}

Motion sensitivity includes a horizontal bias. We estimate optical flow \((u,v)\) via Lucas-Kanade (Eq.\ \eqref{eq:lk}) and modulate flow magnitude by \(\eta(\theta)=1+\kappa|\cos\theta|\) (Eq.\ \eqref{eq:motionbias}) before blending into the image stream.
\begin{equation}
 (u,v)=\arg\min_{u,v}\sum_{y\in W(x)}\big(I_x(y)u+I_y(y)v+I_t(y)\big)^2,\label{eq:lk}
\end{equation}
\begin{equation}
 M(x)=\eta(\theta(x))\,\sqrt{u(x)^2+v(x)^2},\qquad I_{mot}(x)=I_{tmp}(x)+\lambda_M\,\mathrm{norm}(M(x)).\label{eq:motionbias}
\end{equation}

Finally, low-light enhancement consistent with the tapetum lucidum is implemented as a logistic gain with a blue-green tint (Eq.\ \eqref{eq:tapetum}).
\begin{align}
 g(I)&=1+\alpha\,\sigma\big(\beta(\tau-\bar{I})\big),\qquad \bar{I}=\tfrac{1}{3}(R{+}G{+}B),\\
 I_{tap}(x)&=T\big(\,g(I_{mot}(x))\cdot I_{mot}(x)\,\big),\qquad T=\mathrm{diag}(t_R, t_G, t_B),\ t_G\gtrsim t_B\ge t_R,\label{eq:tapetum}
\end{align}
with \(T\) imparting a mild blue-green bias; the final output is \(I_{cat}=I_{tap}\).

We acknowledge certain limitations and present the transformation as an engineering approximation, not a full optical-retinal-cortical simulator. It omits wavelength-dependent blur, detailed retinal sampling mosaics, chromatic aberrations, and dynamic pupil control. As such, it should be interpreted as a near-accurate biologically motivated stressor for invariance analyses rather than a physiologically complete forward model.
We instantiate the filter with the following defaults unless otherwise stated: pupil aspect ratio 3:1 (vertical), rod:cone weighting 25:1, spectral peaks at (450, 556, 498) nm for S/L/rod, spatial acuity reduction equivalent to \(\sim\)1/6 of human high-contrast acuity (behavioral estimates in cats often range 3-9 cpd, with higher optical potential \citep{Clark2013}, field of view \(\sim\)200°\,horizontal/140°\,vertical, temporal sensitivity peaking near 10 Hz with reduced gain beyond 50 Hz, and enhanced weighting of horizontally directed motion.

\subsection{Models, Feature Extraction, Analyses and Statistical Procedures}
All encoders are used with their canonical pretrained weights and are not fine-tuned, in line with frozen-transfer evaluations that emphasize representation quality independent of adaptation. For convolutional encoders, we probe a small set of semantically comparable stages spanning early, middle, and late processing, plus the global pooling stage when available. For transformer encoders, we probe block-wise token representations (ViT) and stage outputs (Swin). To render representations commensurate across families, we standardize activations to vectors: global average pooling for feature maps and class token (or mean over tokens if no class token is defined) for tokenized outputs \citep{dosovitskiy2021imageworth16x16words,liu2021swintransformerhierarchicalvision}. This adheres to common practice in transfer and representational analysis.

Self-supervised models trained with knowledge distillation and masked/teacher-student objectives (e.g., DINO/DINOv2) are included due to their strong frozen-transfer performance and emergent invariances \citep{caron2021emergingpropertiesselfsupervisedvision,oquab2024dinov2learningrobustvisual, siméoni2025dinov3}. For completeness, we include both base and large capacity variants where available to assess whether capacity modulates cross-species alignment. All images are normalized according to each encoder’s canonical evaluation recipe (e.g., ImageNet mean/variance for CNNs; processor-defined normalization for transformers). We batch inputs per domain to ensure identical batch statistics across human and feline images and report per-layer sample counts. Optional ablations operate on a fixed-size, deterministically sampled subset to measure stability. We quantify alignment between human and feline representations at each layer using complementary geometric and statistical tools, and we control for multiple comparisons across models, layers, and metrics. We interleave each method with its mathematical definition and refer to equations  and t-SNE, UMAP visualizations as shown below.
\paragraph{Representational geometry.}
We report Centered Kernel Alignment (CKA) in both its linear form and with an RBF kernel. Linear CKA is invariant to orthogonal transformations and isotropic scaling and has been shown to be stable across architectures and training runs \citep{kornblith2019similarityneuralnetworkrepresentations}. Given paired samples \(X=\{x_i\}_{i=1}^n\), \(Y=\{y_i\}_{i=1}^n\), linear CKA normalizes HSIC between centered Gram matrices (Eq.\ \eqref{eq:cka_lin}).
\begin{equation}
 \mathrm{CKA}_{\text{lin}}(X,Y)=\frac{\operatorname{HSIC}(K,L)}{\sqrt{\operatorname{HSIC}(K,K)\,\operatorname{HSIC}(L,L)}},\quad K=HXX^\top H,\ L=HYY^\top H,\label{eq:cka_lin}
\end{equation}
with \(H=I-\tfrac{1}{n}\mathbf{1}\mathbf{1}^\top\) and \(\operatorname{HSIC}(K,L)=\tfrac{1}{(n-1)^2}\operatorname{tr}(KL)\). RBF CKA replaces \(K,L\) by RBF kernels (Eq.\ \eqref{eq:rbf_cka}).
\begin{equation}
 K_{ij}=\exp\!\Big(-\tfrac{\lVert x_i-x_j\rVert^2}{2\sigma^2}\Big),\quad L_{ij}=\exp\!\Big(-\tfrac{\lVert y_i-y_j\rVert^2}{2\sigma^2}\Big),\quad \mathrm{CKA}_{\text{rbf}}\text{ as in Eq.~}\eqref{eq:cka_lin}.\label{eq:rbf_cka}
\end{equation}
In parallel, we conduct Representational Similarity Analysis (RSA) by computing cosine-based representational dissimilarity matrices (RDMs) per domain (Eq.\ \eqref{eq:rdm_cos}) and correlating the upper triangles using Spearman rank correlation. Significance is assessed with the Mantel permutation test (Eq.\ \eqref{eq:mantel}).
\begin{equation}
 D^X_{ij}=1-\frac{x_i^\top x_j}{\lVert x_i\rVert\,\lVert x_j\rVert},\quad D^Y_{ij}=1-\frac{y_i^\top y_j}{\lVert y_i\rVert\,\lVert y_j\rVert}.\label{eq:rdm_cos}
\end{equation}
Let \(u(\cdot)\) extract upper-triangular entries; the Mantel statistic is \(r=\operatorname{corr}(u(D^X),u(D^Y))\), with p-values from permutations of indices (Eq.\ \eqref{eq:mantel}).
\begin{equation}
 r=\frac{\sum_k (u(D^X)_k-\bar u_X)(u(D^Y)_k-\bar u_Y)}{\sqrt{\sum_k(u(D^X)_k-\bar u_X)^2\,\sum_k(u(D^Y)_k-\bar u_Y)^2}},\ \ p\text{-value by }\pi\text{-permutations}.\label{eq:mantel}
\end{equation}

\paragraph{Distributional shift tests.}
To detect shifts beyond geometric alignment, we compute the Maximum Mean Discrepancy (MMD; RBF kernel), the Energy distance, and a projected 1-Wasserstein distance. For MMD we use the unbiased estimator with an RBF kernel (Eq.\ \eqref{eq:mmd}); p-values are obtained via label permutations. Energy distance is computed from pairwise Euclidean distances (Eq.\ \eqref{eq:energy}). For interpretability, we project onto the first principal component and compute the 1D \(W_1\) between the projected empirical distributions (Eq.\ \eqref{eq:w1}).
\begin{align}
 \widehat{\mathrm{MMD}}^2 &= \frac{1}{m(m-1)}\sum_{i\ne j} k(x_i,x_j)+\frac{1}{n(n-1)}\sum_{i\ne j} k(y_i,y_j)-\frac{2}{mn}\sum_{i=1}^m\sum_{j=1}^n k(x_i,y_j),\label{eq:mmd}\\
 k(u,v)&=\exp\!\Big(-\tfrac{\lVert u-v\rVert^2}{2\sigma^2}\Big).\nonumber
\end{align}
\begin{equation}
 \widehat{\mathcal{E}}^2=\frac{2}{mn}\sum_{i,j}\lVert x_i-y_j\rVert_2-\frac{1}{m^2}\sum_{i,j}\lVert x_i-x_j\rVert_2-\frac{1}{n^2}\sum_{i,j}\lVert y_i-y_j\rVert_2.\label{eq:energy}
\end{equation}
\begin{equation}
 W_1=\frac{1}{K}\sum_{k=1}^K \big| s_{(k)}-t_{(k)}\big|,\quad s_i=w^\top x_i,\ t_j=w^\top y_j,\ w=\text{first PC}.\label{eq:w1}
\end{equation}

\paragraph{Paired similarity.}
For each layer we compute per-pair cosine similarity and Euclidean distance between human and feline vectors. To test for mean shifts, we project differences onto the first principal component \(w\) of \(\{x_i-y_i\}\) and test \(H_0:\mathbb{E}[d]=0\) on \(d_i=w^\top(x_i-y_i)\). If Shapiro-Wilk fails to reject normality, we report the paired t-statistic (Eq.\ \eqref{eq:tstat}); otherwise we report the Wilcoxon signed-rank statistic.
\begin{equation}
 t=\frac{\bar d}{s_d/\sqrt{n}},\quad \bar d=\frac{1}{n}\sum_i d_i,\ s_d^2=\frac{1}{n-1}\sum_i (d_i-\bar d)^2.\label{eq:tstat}
\end{equation}

\paragraph{Multiple testing.}
We aggregate all collected p-values and apply the Benjamini-Hochberg procedure (FDR level 0.05) to obtain q-values and rejection decisions (Eq.\ \eqref{eq:bh}). This controls false discoveries across the large grid of model-layer-metric combinations. We also report q-values via the standard step-up estimator.
\begin{equation}
 k=\max\Big\{i:\ p_{(i)}\le \tfrac{i}{M}\,q\Big\},\qquad \text{reject } H_{(1)},\ldots,H_{(k)},\quad q_{(i)}=\min_{j\ge i}\frac{M}{j}\,p_{(j)}.\label{eq:bh}
\end{equation}

We employ qualitative and quantitative visualization to contextualize the statistical findings and to diagnose failure modes.

First, we project high-dimensional layer-wise representations to two dimensions using t-SNE \citep{vandermaaten08a} and UMAP \cite{McInnes2018}. For each model and layer, we co-embed human and feline vectors and inspect whether clusters corresponding to domains separate or overlap. t-SNE perplexity is adapted to the sample size, and UMAP neighborhoods scale with data density; these settings provide stable 2D summaries while avoiding over-interpretation of local distances.

Second, we visualize intermediate activations for paired images. For convolutional layers, we render channel-wise activation maps after min-max normalization; for transformer blocks, we compute token saliency maps by averaging over the embedding dimension and reshaping to the token grid (dropping class tokens when appropriate). We display human and feline activations side-by-side to highlight convergences and divergences at the same layer.

These visualizations serve as diagnostic aids rather than formal tests: they illustrate patterns suggested by the alignment metrics (e.g., strong overlap in 2D projections when CKA is high) and reveal layer-wise phenomena such as texture vs. shape preferences or spatial pooling differences that may underlie quantitative effects.

\captionsetup{font=small,labelfont=bf}

\begin{figure*}[t]
  \centering
  \includegraphics[width=\textwidth]{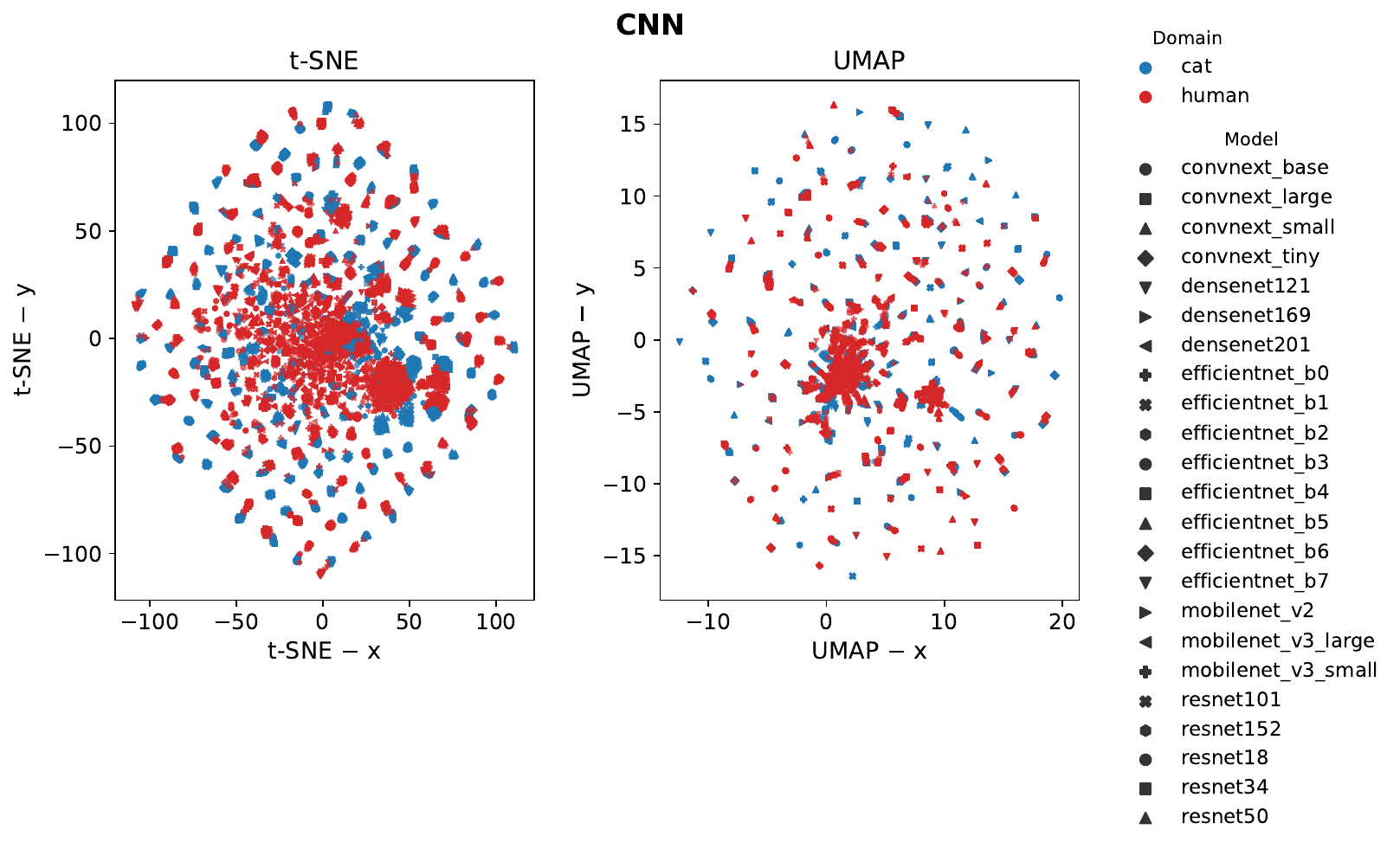}
  \caption{CNN embeddings with t-SNE (left) and UMAP (right). Colors encode domains (human vs. cat) and marker shapes encode models within the family. These panels are intended to assess domain-level overlap by visual inspection: color mixing indicates cross-domain similarity, while separated color clusters indicate stronger domain-specific structure; shape differences reveal whether such trends are consistent across CNN variants.}
  \label{fig:cnn-embeddings}
\end{figure*}

\begin{figure*}[t]
  \centering
  \includegraphics[width=\textwidth]{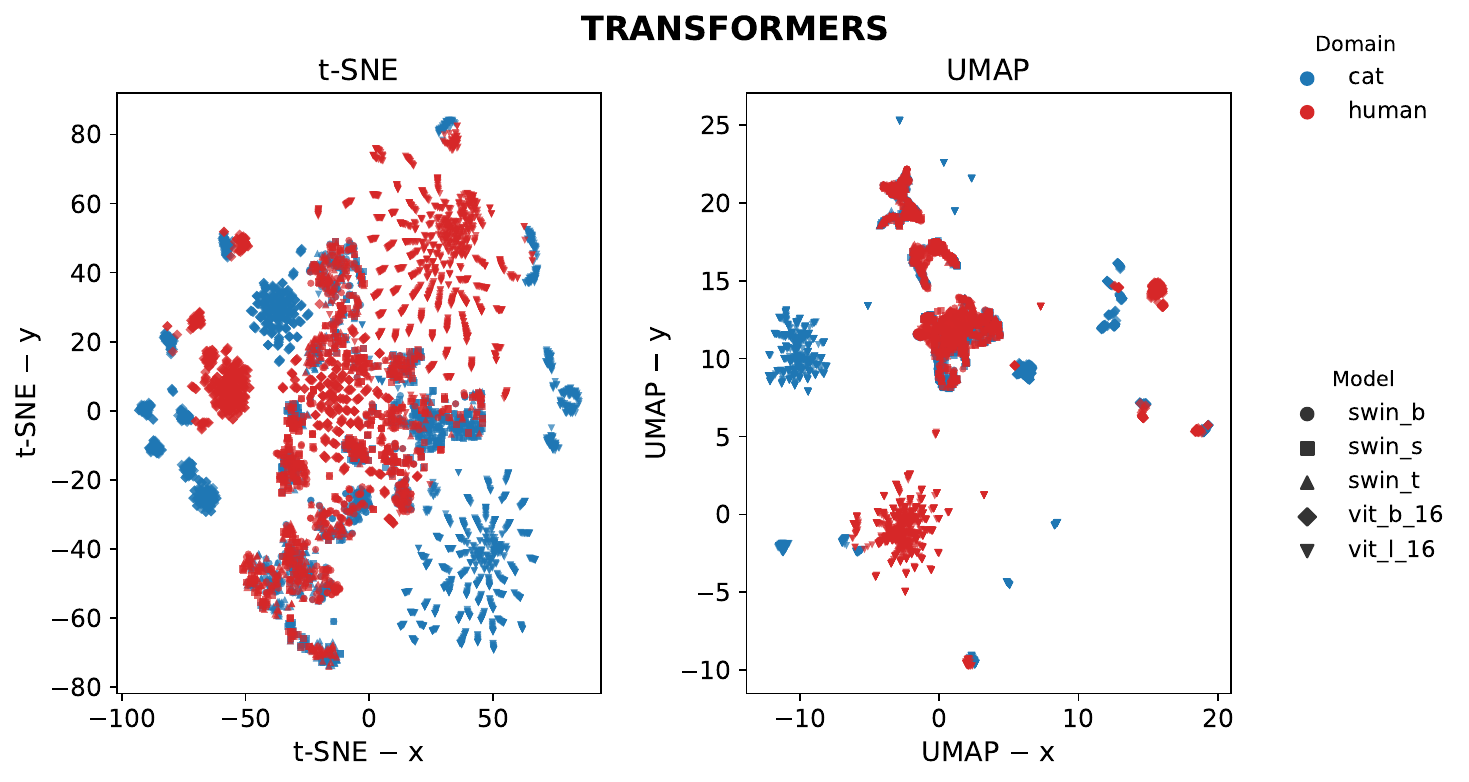}
  \caption{Transformer embeddings with t-SNE (left) and UMAP (right). Colors (domains) and marker shapes (models) follow Figure~\ref{fig:cnn-embeddings}. Showing both t-SNE and UMAP allows a robustness check: consistent patterns across methods lend confidence, while differences may reflect method-specific neighborhood preservation.}
  \label{fig:transformers-embeddings}
\end{figure*}

\begin{figure*}[!htbp]
  \centering
  \includegraphics[width=\textwidth]{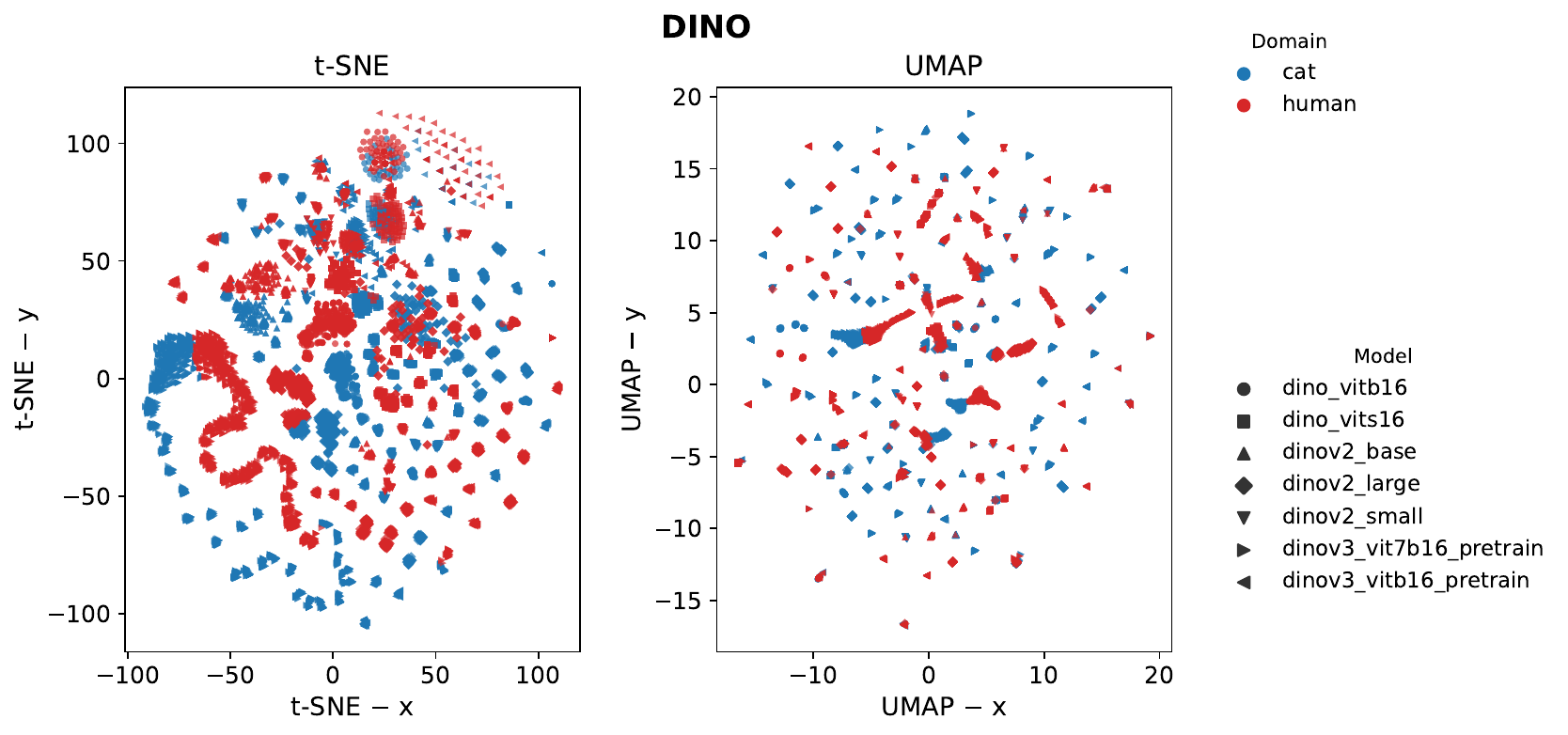}
  \caption{DINO embeddings with t-SNE (left) and UMAP (right). Colors denote domains; marker shapes denote DINO variants. Self-supervised representations often yield distinct geometry; these panels enable visual examination of domain separation vs. overlap and whether patterns are consistent across DINO variants.}
  \label{fig:dino-embeddings}
\end{figure*}

A brief summary table of reported metrics and statistical controls is provided in Table \ref{tab:metrics_controls}.
\begin{table}[t]
\centering
\caption{Summary of reported metrics and statistical controls.}
\label{tab:metrics_controls}
\setlength{\tabcolsep}{6pt}
\renewcommand{\arraystretch}{1.15}
\begin{tabularx}{\linewidth}{@{}%
  >{\RaggedRight\arraybackslash}p{2.9cm}%
  >{\RaggedRight\arraybackslash}p{6.0cm}%
  >{\RaggedRight\arraybackslash}Y@{}}
\toprule
Report & Contents & Notes \\
\midrule
Layer-wise alignment & CKA (linear, RBF); RSA Spearman; Mantel \(r/p\) & 500 permutations for Mantel \\
Distribution shift & MMD statistic/\(p\); Energy statistic/\(p\); 1-Wasserstein & Median-heuristic bandwidth for MMD \\
Paired stability & Mean cosine, mean \(L_2\); test statistic/\(p\); test type & Shapiro-Wilk for test selection \\
Multiple testing & Raw \(p\), \(q\)-value, rejection & Benjamini-Hochberg at 0.05 \\
\bottomrule
\end{tabularx}
\end{table}
\section{Results}
We evaluate cross-species representational alignment on frozen encoders across three families: CNNs (ResNet, MobileNet, DenseNet, EfficientNet, ConvNeXt), supervised transformers (ViT, Swin), and self-supervised transformers (DINO, DINOv2/v3). Following the methodology, we report layer-wise alignment aggregated per model via CKA (linear and RBF), RSA/Mantel, and distribution-shift and paired metrics, and we identify the best-performing layer or block by CKA.

Table~\ref{tab:results_overall} summarizes model-level aggregates from the three familes: CNNs, Transformers(ViTs) and DINOv1/v2/v3. We use mean RBF-CKA as the primary selection criterion, as it is sensitive to certain nonlinear correspondences while remaining stable in practice; linear-CKA and RSA trends are consistent.

Across families, the self-supervised Vision Transformer \textbf{DINO ViT-B/16} achieves the highest mean RBF-CKA (\textbf{0.8144}), closely followed by supervised \textbf{ViT-L/16} (\textbf{0.8057}). Among CNNs, \textbf{EfficientNet-B3} yields the strongest mean RBF-CKA (\textbf{0.7017}). These results indicate that transformer-based encoders, particularly self-supervised ViTs, preserve stronger cross-species invariances under our paired design.

\begin{table}[t]
\centering
\caption{Model-level aggregates from overall summaries. For selection we prioritize mean RBF-CKA; best within each family and overall are bolded. Values are taken from the three overall summary CSVs.}
\label{tab:results_overall}
\setlength{\tabcolsep}{4pt}
\renewcommand{\arraystretch}{1.15}
\begin{tabularx}{\linewidth}{@{}%
  >{\RaggedRight\arraybackslash}p{1.9cm}%
  >{\RaggedRight\arraybackslash}p{3.2cm}%
  >{\RaggedRight\arraybackslash}p{2.4cm}%
  >{\RaggedRight\arraybackslash}Y%
  >{\RaggedRight\arraybackslash}Y%
  >{\RaggedRight\arraybackslash}Y%
  >{\RaggedRight\arraybackslash}Y@{}}
\toprule
Family & Model & Best layer/block & CKA-RBF (mean) & CKA-Linear (mean) & RSA (mean) & Mean cosine \\
\midrule
CNN & EfficientNet-B3 & stage5 & \textbf{0.7017} & 0.6371 & 0.5344 & 0.6308 \\
CNN & ResNet-50 & layer3 & 0.6902 & 0.6628 & 0.4876 & 0.6022 \\
CNN & DenseNet-169 & db3 & 0.6853 & 0.6166 & 0.5417 & 0.7036 \\
CNN & EfficientNet-B1 & stage5 & 0.6838 & 0.6389 & 0.5107 & 0.4939 \\
CNN & ConvNeXt-L & stage1 & 0.5599 & 0.5355 & 0.5428 & 0.8292 \\
\midrule
Transformer (sup.) & ViT-L/16 & block14 & \textbf{0.8057} & 0.7050 & 0.4647 & 0.5960 \\
Transformer (sup.) & ViT-B/16 & block8 & 0.7755 & 0.6840 & 0.5266 & 0.6943 \\
Transformer (sup.) & Swin-B & stage3 & 0.4688 & 0.4269 & 0.3818 & 0.6110 \\
\midrule
Self-sup. (DINO) & DINO ViT-B/16 & block0 & \textbf{0.8144} & 0.7446 & 0.6980 & 0.7995 \\
Self-sup. (DINO) & DINO ViT-S/16 & block0 & 0.7682 & 0.6991 & 0.6668 & 0.8384 \\
Self-sup. (DINOv2) & DINOv2-Base & block0 & 0.7232 & 0.6082 & 0.5669 & 0.8454 \\
\midrule
Overall best & \textbf{DINO ViT-B/16} & block0 & \textbf{0.8144} & 0.7446 & 0.6980 & 0.7995 \\
\bottomrule
\end{tabularx}
\end{table}
We make the following family-specific observations:
\begin{enumerate}
    \item \textbf{CNNs.} EfficientNet variants perform strongly, with B3 (CKA-RBF mean 0.7017) leading, followed closely by ResNet-50 (0.6902) and DenseNet-169 (0.6853). Best alignment typically occurs at later blocks (e.g., EfficientNet-B3 stage5; ResNet-50 layer3), consistent with hierarchical convergence.
    \item \textbf{Supervised transformers.} ViT-L/16 (0.8057) outperforms Swin variants by a wide margin; best alignment arises at deeper transformer blocks (block14 for ViT-L/16; block8 for ViT-B/16).
    \item \textbf{Self-supervised transformers.} DINO ViT-B/16 achieves the highest overall alignment (0.8144); DINOv2-Base is strong (0.7232), while DINOv3 pretrain variants show moderate alignment in our setting.
\end{enumerate}
Mantel permutation tests, MMD, and Energy distance frequently reject the null across layers (see summary CSV counts), confirming measurable distributional differences even when alignment is high. Nevertheless, the leading models maintain robust cross-domain alignment by CKA and RSA, suggesting shape/semantic consistency despite domain shifts.

Complete per-model tables for each family, including all reported aggregates are provided in Tables~\ref{tab:cnn_aggregates}, ~\ref{tab:transformers_supervised_aggregates}, ~\ref{tab:selfsup_transformers_aggregates} in Appendix~\ref{app:cnn}. We also investigate layers with most dissimilarities. See Tables~\ref{tab:cnn_dissimilar}, ~\ref{tab:transformer_dissimilar}, ~\ref{tab:dino_dissimilar} in Appendix~\ref{app:layer_dissimilarity}
\captionsetup{font=small,labelfont=bf}
Beyond alignment, we systematically localize layers with strongest dissimilarity signals using low CKA/RSA and high projected 1D Wasserstein. Three robust patterns emerge across families: (i) \emph{lowest alignment concentrates early}: initial CNN convolutions (e.g., ResNet/conv1) and the earliest blocks in ViT/DINO variants tend to have the lowest CKA-Linear and RSA; (ii) \emph{distributional shift can peak late}: deeper EfficientNet stages and late ViT blocks exhibit the largest Wasserstein shifts while still retaining moderate alignment; (iii) \emph{self-supervised giants can decouple geometry and distribution}: DINOv3 large models show very high late-block Wasserstein despite competitive CKA/RSA. Taken together, early layers are the most geometry-dissimilar, whereas certain deep layers carry the largest magnitude/distribution differences.
\begin{figure*}[t]
  \centering
  \includegraphics[width=\textwidth]{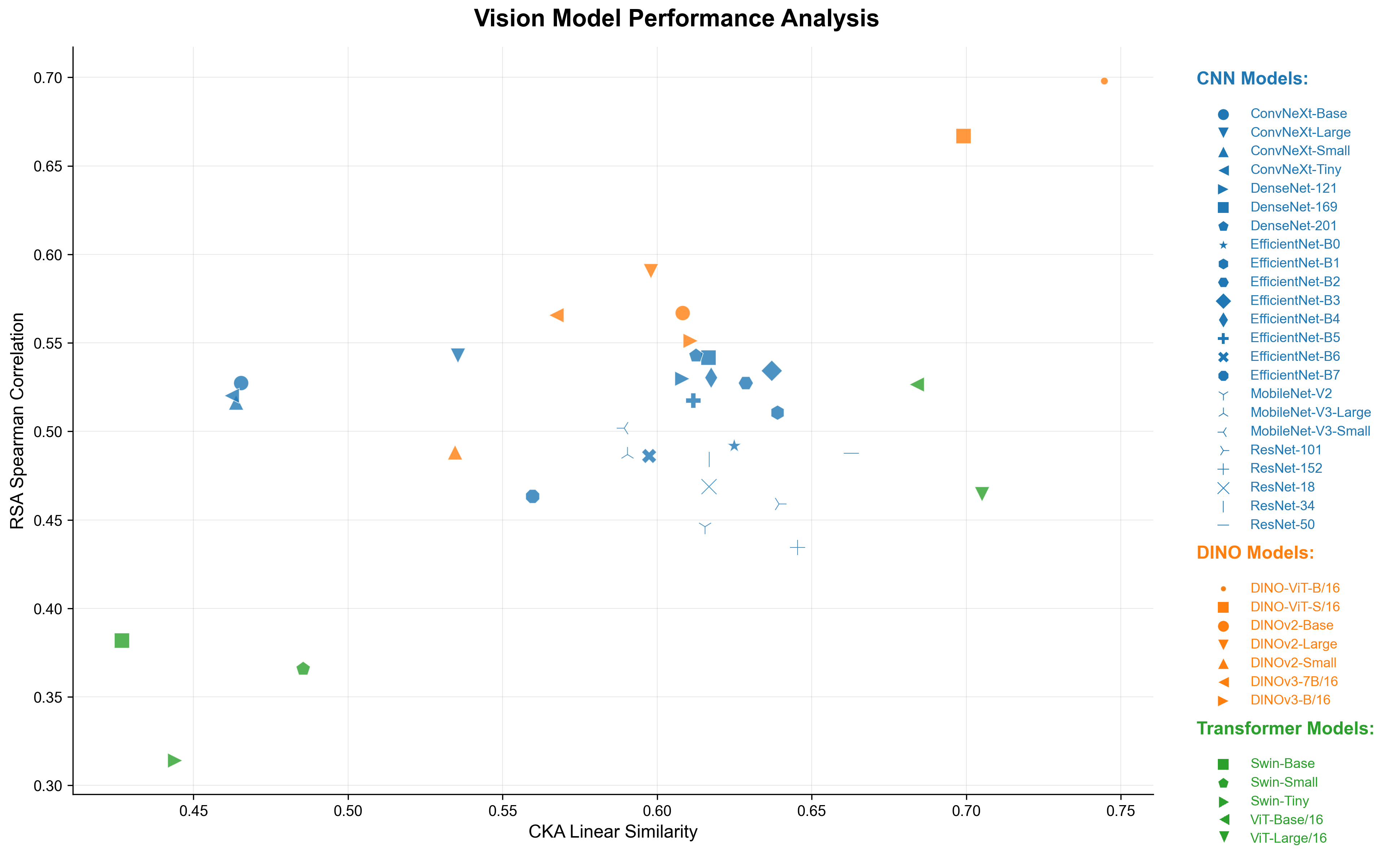}
  \caption{\textbf{DINO} models cluster in the upper-right region (high performance on both metrics), with DINO-ViT-B/16 achieving the highest RSA Spearman performance (0.698). \textbf{CNN} models form a tight cluster in the middle region, with EfficientNet variants generally performing better than ResNet and DenseNet models on both metrics. \textbf{Transformer} models show high variability, with ViT models (ViT-B/16, ViT-L/16) achieving high CKA Linear performance but moderate RSA Spearman scores, while Swin transformers cluster in the lower-left region.}
  \label{fig:vision_models_analysis}
\end{figure*}

\begin{figure*}[t]
  \centering
  \includegraphics[width=\textwidth]{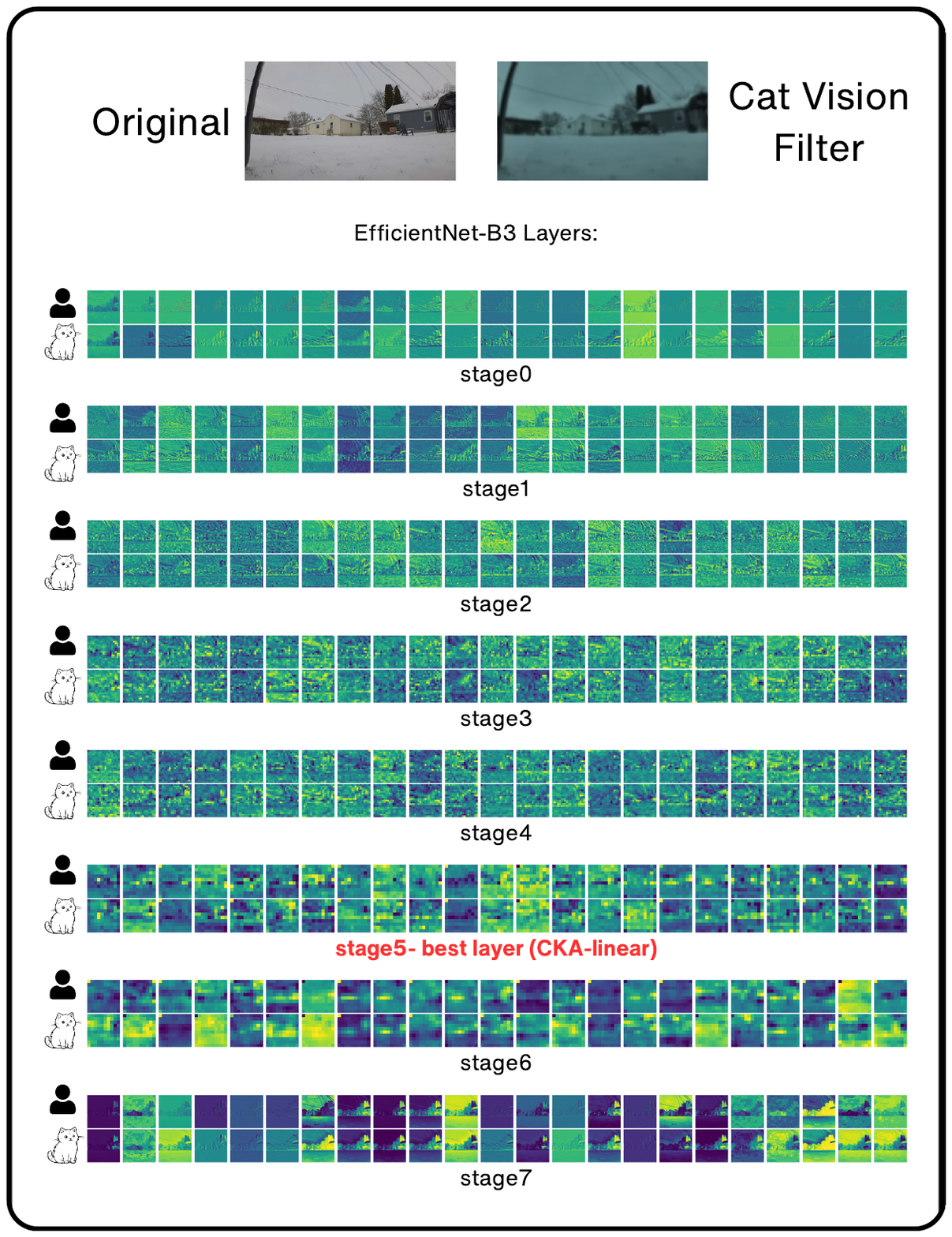}
  \caption{Layer-wise feature maps for the best performing model in CNN family; EfficientNet-B3 with best layer by CKA-linear: stage5}
  \label{fig:cnn_best}
\end{figure*}

\begin{figure*}[t]
  \centering
  \includegraphics[width=\textwidth]{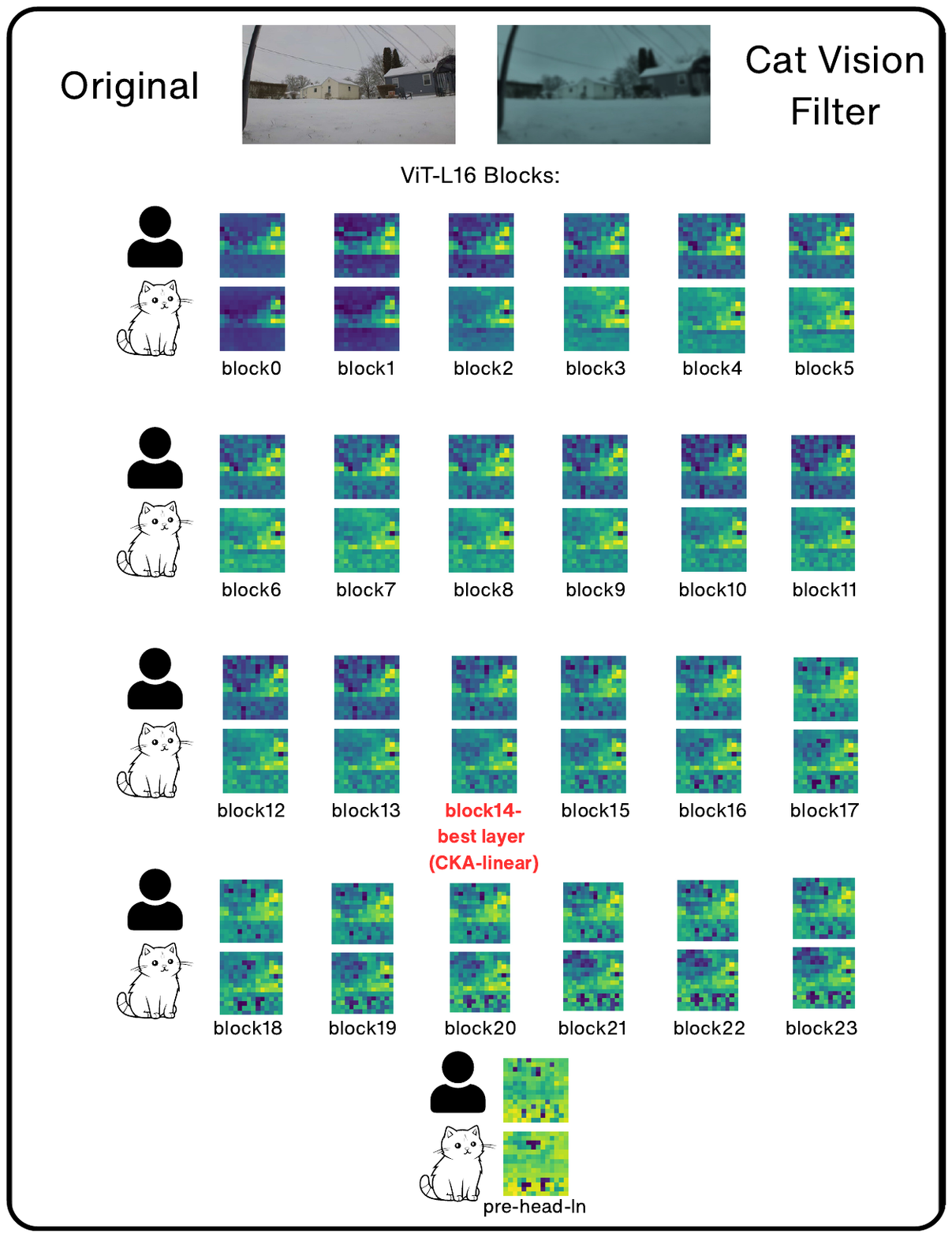}
  \caption{Block-wise visualization for the best performing model in Transformers family; ViT l16; best block by CKA-linear: block14}
  \label{fig:transformers_best}
\end{figure*}

\begin{figure*}[t]
  \centering
  \includegraphics[width=\textwidth]{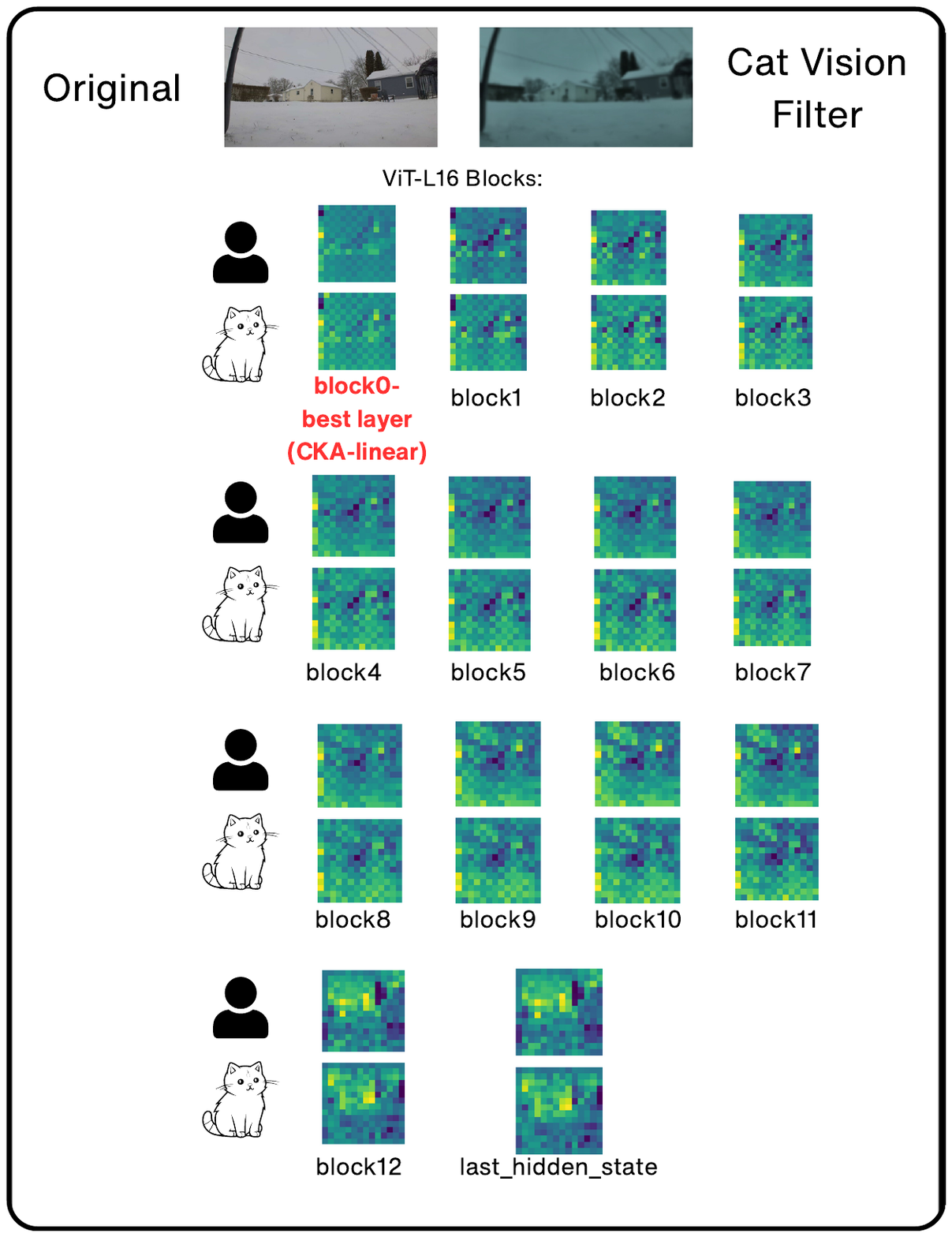}
  \caption{Block-wise visualization for the best performing model in DINO family; DINO ViT b16; best block by CKA-linear: block0}
  \label{fig:dino_best}
\end{figure*}

We provide Payer-wise feature maps for all three model families as seen in ~\ref{fig:cnn_best}, ~\ref{fig:transformers_best} and ~\ref{fig:dino_best}.

\section{Discussion}
We investigated cross-species representational invariance using frozen encoders on strictly paired human-feline views, triangulating alignment (CKA linear/RBF, RSA/Mantel), distributional shift (MMD, Energy, projected 1D Wasserstein), and paired similarity (cosine, L2) to characterize architectural behavior across domains. We summarize findings, likely causes, implications, scope choices, and future directions.

Transformers, especially self-supervised ViTs show the strongest alignment: DINO ViT-B/16 has the highest mean RBF-CKA (0.8144), followed by supervised ViT-L/16 (0.8057); the top CNN is EfficientNet-B3 (0.7017). Linear CKA means mirror this (DINO ViT-B/16: 0.7446; ViT-L/16: 0.7050; EfficientNet-B3: 0.6371), as do RSA/Mantel (0.698; 0.465; 0.534). Three factors plausibly drive these trends: (i) token-based global attention supports shape and part-whole consistency; (ii) invariance-focused self-supervision (DINO/DINOv2/v3) avoids human-label biases; and (iii) capacity/depth (e.g., ViT-L/16) enables abstractions aligning semantic manifolds. DINO ViT-B/16’s RBF-CKA edge over supervised ViTs underscores the role of the objective beyond architecture.

Best layers differ by family. Supervised ViTs peak deeper (ViT-L/16 at block14; ViT-B/16 at block8). DINO often peaks at block0 for linear CKA, suggesting early, domain-general structure under our preprocessing. CNNs peak late (EfficientNet-B3 at stage5; ResNet-50 at layer3), consistent with hierarchical abstraction.

Despite high CKA/RSA, shift tests frequently reject after BH-FDR. CNNs show large Energy means (EfficientNet-B3: 20.0463) and projected 1D Wasserstein (EfficientNet-B3: 26.6422); supervised ViTs show moderate projected Wasserstein (ViT-B/16: 6.4436; ViT-L/16: 34.3499); DINO ranges widely (DINO ViT-B/16: 11.1702; DINOv3 ViT-7B/16 pretrain: 263.1593). MMD/Energy rejections are common (CNNs: 7-9; supervised ViTs: up to 25; DINO/DINOv3: teens or higher). Geometry metrics (CKA/RSA) are invariant to uniform scaling/affine transforms, whereas MMD/Energy/Wasserstein probe absolute distribution; cosine is magnitude-invariant, L2 is not. ConvNeXt variants exemplify this (convnext\_large mean cosine: 0.8292 with larger L2), indicating similar directions but norm shifts, likely from domain-specific factors (e.g., illumination).

EfficientNet-B3 is the top CNN by mean RBF-CKA (0.7017), edging ResNet-50 (0.6902) and DenseNet-169 (0.6853); late-layer alignment (stage5/layer3) is typical. CNNs often have larger Energy/Wasserstein, consistent with mass/magnitude shifts. MobileNet variants are competitive for size but trail EfficientNet/ResNet. ViT-L/16 (0.8057) surpasses ViT-B/16 (0.7755) and Swin (\(\sim\)0.46-0.50). Best alignment at deeper blocks (block14 for ViT-L/16) supports semantics-driven consistency. Swin’s windowed attention is efficient but less aligned cross-species than global ViTs here. DINO ViT-B/16 (0.8144) leads overall; DINOv2-Base (0.7232) is competitive; DINOv3 pretrains show moderate alignment but sometimes very large shift statistics (e.g., projected Wasserstein 263.1593 for ViT-7B/16 pretrain), indicating substantial distributional differences despite geometric agreement.

Three implications follow. First, training objective matters: self-supervised ViTs consistently top mean RBF-CKA (e.g., DINO ViT-B/16), yielding cross-species alignment without labels. Second, hierarchy matters: deeper blocks/stages align best, guiding feature extraction for cross-domain stability. Third, geometry and distribution are complementary: prioritize geometric alignment, then apply lightweight calibration (feature normalization, small adapters) to reduce residual MMD/Energy/Wasserstein shifts without altering geometry.

We freeze all encoders and use strictly paired human-feline views to isolate architectural/pretraining biases, report aggregates (means/medians; best layers/blocks) for stability, and include projected 1D Wasserstein alongside MMD/Energy for interpretability. These choices localize where alignment arises along the hierarchies. Together, these results indicate that training objective, architectural hierarchy, and capacity jointly shape cross-species representational invariance, and that aligning geometry and distribution requires complementary tools.

A primary reason for investigation of the feline visual system was the apparent difference in the retinal structure between humans and cats and also the initial works by Hubel and Wiesel on cat neuronal fields that ushered in the study of the visual system \citep{HubelWiesel1959}.  Neuroscientifically one would assume that given the differences in the retinal structure, gray matter volume, brain folding patterns, layer differences amongst animal and human models \citep{Krubitzer2009,Yin2025.03.05.641692}, condensed brain hierarchy in small mammalian models \citep{Lanfranchi2025}, we would see a striking difference in cat representation of the world but apparently that's not the case. The hierarchical organization of the visual system results in invariant and stable representations of the perceived world which highlights the convergent evolutionary characteristics of the visual model. Future work can build directly on this foundation in several directions that are complementary to our goals here. (i) \emph{Species generalization.} Extending the paired protocol to additional animals (e.g., canine, nonhuman primate, avian) would test how architectural and objective choices scale with ecological and optical diversity and evolutionary goals; this creates opportunities to connect to systems neuroscience by comparing model RDMs to neural data across species and cortical hierarchies. (ii) \emph{Visual deficiencies.} We can use such an approach to investigate visual deficiences like partial blindness, myopia and hypermetropia and how corrections at different stages of visual development can impact visual processing and detection.(iii) \emph{Non-visual sensory modalities.} This line of work could be extended to incorporate differences in auditory, tactile and olfactory senses across animal models.

\section*{Acknowledgments}
We acknowledge National Supercomputing 
Mission (NSM) for providing computing resources of ‘PARAM Ananta’ at IIT Gandhinagar, 
which is implemented by C-DAC and supported by the Ministry of Electronics and 
Information Technology (MeitY) and Department of Science and Technology (DST), 
Government of India.

\bibliographystyle{unsrt}  
\bibliography{references}  

\appendix
\section{Appendix A. Per Model Summary Tables}
\label{app:cnn}
\begin{table}[!htbp]
\centering
\caption{EfficientNet-B3 achieved the highest mean RBF-CKA among CNNs (0.7017), with best alignment at stage5.}
\label{tab:cnn_aggregates}
\setlength{\tabcolsep}{4pt}
\renewcommand{\arraystretch}{1.12}
\begin{tabularx}{\linewidth}{@{}%
  >{\RaggedRight\arraybackslash}p{2.9cm}%
  >{\RaggedRight\arraybackslash}p{2.2cm}%
  >{\RaggedRight\arraybackslash}Y%
  >{\RaggedRight\arraybackslash}Y%
  >{\RaggedRight\arraybackslash}Y%
  >{\RaggedRight\arraybackslash}Y%
  >{\RaggedRight\arraybackslash}Y%
  >{\RaggedRight\arraybackslash}Y@{}}
\toprule
Model & Best layer & CKA-RBF mean & CKA-RBF max & CKA-Linear mean & RSA mean & Mean cosine & Mean L2 \\
\midrule
convnext\_base & stage1 & 0.5212 & 0.6015 & 0.4653 & 0.5274 & 0.8612 & 1.3488 \\
convnext\_large & stage1 & 0.5599 & 0.6743 & 0.5355 & 0.5428 & 0.8292 & 2.1653 \\
convnext\_small & stage2 & 0.5142 & 0.6178 & 0.4638 & 0.5163 & 0.8108 & 0.8129 \\
convnext\_tiny & stage1 & 0.5473 & 0.6780 & 0.4624 & 0.5202 & 0.7473 & 0.9895 \\
DenseNet-121 & tr2 & 0.6776 & 0.8606 & 0.6081 & 0.5299 & 0.7090 & 4.6149 \\
DenseNet-169 & db3 & 0.6853 & 0.8721 & 0.6166 & 0.5417 & 0.7036 & 4.8734 \\
DenseNet-201 & tr3 & 0.6825 & 0.8870 & 0.6125 & 0.5429 & 0.7132 & 4.8342 \\
EfficientNet-B0 & stage4 & 0.6627 & 0.8560 & 0.6249 & 0.4920 & 0.4418 & 19.6399 \\
EfficientNet-B1 & stage5 & 0.6838 & 0.8813 & 0.6389 & 0.5107 & 0.4939 & 21.9126 \\
EfficientNet-B2 & stage4 & 0.6706 & 0.8743 & 0.6287 & 0.5273 & 0.5645 & 20.2296 \\
EfficientNet-B3 & stage5 & 0.7017 & 0.8743 & 0.6371 & 0.5344 & 0.6308 & 26.6422 \\
EfficientNet-B4 & stage5 & 0.6625 & 0.8591 & 0.6175 & 0.5304 & 0.6339 & 44.0907 \\
EfficientNet-B5 & stage3 & 0.6794 & 0.8393 & 0.6117 & 0.5175 & 0.7026 & 20.0516 \\
EfficientNet-B6 & stage6 & 0.6607 & 0.8553 & 0.5974 & 0.4862 & 0.7057 & 20.0586 \\
EfficientNet-B7 & stage3 & 0.6292 & 0.7974 & 0.5597 & 0.4634 & 0.7156 & 21.8801 \\
mobilenet\_v2 & mid3 & 0.6517 & 0.8263 & 0.6154 & 0.4463 & 0.2946 & 16.1492 \\
mobilenet\_v3\_large & final & 0.6198 & 0.8597 & 0.5903 & 0.4871 & 0.3545 & 12.4279 \\
mobilenet\_v3\_small & mid3 & 0.6664 & 0.8369 & 0.5893 & 0.5018 & 0.4325 & 4.4553 \\
ResNet-18 & layer2 & 0.6759 & 0.8370 & 0.6167 & 0.4689 & 0.7260 & 8.2918 \\
ResNet-34 & layer3 & 0.6730 & 0.8609 & 0.6169 & 0.4841 & 0.7324 & 9.1681 \\
ResNet-50 & layer3 & 0.6902 & 0.8988 & 0.6628 & 0.4876 & 0.6022 & 12.8899 \\
ResNet-101 & layer3 & 0.6787 & 0.9055 & 0.6394 & 0.4591 & 0.6153 & 16.1519 \\
ResNet-152 & layer3 & 0.6868 & 0.9081 & 0.6455 & 0.4344 & 0.6253 & 17.8186 \\
\bottomrule
\end{tabularx}
\end{table}

\label{app:transformers}
\begin{table}[!htbp]
\centering
\caption{ViT-L/16 achieved the highest mean RBF-CKA among supervised transformers (0.8057), with best alignment at block14.}
\begin{description}
    \item[] 
\end{description}
\label{tab:transformers_supervised_aggregates}
\setlength{\tabcolsep}{4pt}
\renewcommand{\arraystretch}{1.12}
\begin{tabularx}{\linewidth}{@{}%
  >{\RaggedRight\arraybackslash}p{2.8cm}%
  >{\RaggedRight\arraybackslash}p{2.6cm}%
  >{\RaggedRight\arraybackslash}Y%
  >{\RaggedRight\arraybackslash}Y%
  >{\RaggedRight\arraybackslash}Y%
  >{\RaggedRight\arraybackslash}Y%
  >{\RaggedRight\arraybackslash}Y%
  >{\RaggedRight\arraybackslash}Y@{}}
\toprule
Model & Best block/stage & CKA-RBF mean & CKA-RBF max & CKA-Linear mean & RSA mean & Mean cosine & Mean L2 \\
\midrule
Swin-T & stage3 & 0.4624 & 0.5789 & 0.4440 & 0.3142 & 0.6217 & 0.4581 \\
Swin-S & stage3 & 0.4993 & 0.5802 & 0.4854 & 0.3660 & 0.6723 & 0.5027 \\
Swin-B & stage3 & 0.4688 & 0.6038 & 0.4269 & 0.3818 & 0.6110 & 0.3683 \\
ViT-B/16 & block8 & 0.7755 & 0.9251 & 0.6840 & 0.5266 & 0.6943 & 6.4436 \\
ViT-L/16 & block14 & 0.8057 & 0.9338 & 0.7050 & 0.4647 & 0.5960 & 34.3499 \\
\bottomrule
\end{tabularx}
\end{table}
\label{app:dino}
\begin{table}[!htbp]
\centering
\caption{DINO ViT-B/16 achieved the highest mean RBF-CKA among self-supervised transformers (0.8144), with best alignment at block0.}
\label{tab:selfsup_transformers_aggregates}
\setlength{\tabcolsep}{4pt}
\renewcommand{\arraystretch}{1.12}
\begin{tabularx}{\linewidth}{@{}%
  >{\RaggedRight\arraybackslash}p{4.0cm}%
  >{\RaggedRight\arraybackslash}p{2.0cm}%
  >{\RaggedRight\arraybackslash}Y%
  >{\RaggedRight\arraybackslash}Y%
  >{\RaggedRight\arraybackslash}Y%
  >{\RaggedRight\arraybackslash}Y%
  >{\RaggedRight\arraybackslash}Y%
  >{\RaggedRight\arraybackslash}Y@{}}
\toprule
Model & Best block & CKA-RBF mean & CKA-RBF max & CKA-Linear mean & RSA mean & Mean cosine & Mean L2 \\
\midrule
DINO ViT-B/16 & block0 & 0.8144 & 1.0000 & 0.7446 & 0.6980 & 0.7995 & 11.1702 \\
DINO ViT-S/16 & block0 & 0.7682 & 1.0000 & 0.6991 & 0.6668 & 0.8384 & 7.7151 \\
DINOv2-Base & block0 & 0.7232 & 1.0000 & 0.6082 & 0.5669 & 0.8454 & 8.7336 \\
DINOv2-Large & block0 & 0.7029 & 1.0000 & 0.5980 & 0.5906 & 0.8435 & 8.3610 \\
DINOv2-Small & block0 & 0.6464 & 1.0000 & 0.5346 & 0.4881 & 0.9004 & 5.3112 \\
DINOv3 ViT-B/16 (pretrain) & block0 & 0.7092 & 1.0000 & 0.6107 & 0.5513 & 0.8552 & 43.1978 \\
DINOv3 ViT-7B/16 (pretrain) & block0 & 0.6969 & 1.0000 & 0.5673 & 0.5658 & 0.9360 & 263.1593 \\
\bottomrule
\end{tabularx}
\end{table}
\section{Appendix B. Layers with Most Dissimilarity Across Families}
\label{app:layer_dissimilarity}

We report layers/blocks exhibiting the strongest cross-domain dissimilarities, grounded in the per-layer analyses:
(i) lowest alignment by CKA-Linear and RSA Spearman; and (ii) largest distributional shift by projected 1D Wasserstein (with MMD/Energy concordant).
\begin{table}[!htbp]
\centering
\small
\begin{tabular}{llll}
\toprule
Criterion & Model.Layer & Value\\
\midrule
Lowest CKA-Linear & efficientnet\_b3.stage1 & 0.2448\\
Lowest CKA-Linear & densenet201.conv0 & 0.2451\\
Lowest CKA-Linear & efficientnet\_b7.stage1 & 0.2614\\
Lowest CKA-Linear & efficientnet\_b2.stage1 & 0.2694\\
Lowest RSA Spearman & resnet34.conv1 & 0.0750\\
Lowest RSA Spearman & resnet18.conv1 & 0.0759\\
Lowest RSA Spearman & resnet50.conv1 & 0.1013\\
Largest W1 & efficientnet\_b4.stage4 & 74.662\\
Next largest W1 & efficientnet\_b4.stage6 & 58.455\\
Next largest W1 & efficientnet\_b3.stage6 & 30.629\\
\bottomrule
\end{tabular}
\caption{CNN layers with lowest alignment and highest shift.}
\label{tab:cnn_dissimilar}
\end{table}

\begin{table}[!htbp]
\centering
\small
\begin{tabular}{llll}
\toprule
Criterion & Model.Layer & Value\\
\midrule
Lowest CKA-Linear & swin\_b.stage7 & 0.0660\\
Lowest CKA-Linear & swin\_t.stage2 & 0.3724\\
Lowest CKA-Linear & vit\_l\_16.block22 & 0.2814\\
Lowest RSA Spearman & swin\_b.norm & 0.0636\\
Lowest RSA Spearman & swin\_t.stage2 & 0.0922\\
Lowest RSA Spearman & swin\_b.stage7 & 0.0990\\
Largest W1 & vit\_l\_16.block21-23 & \(\approx\)30.25-30.42\\
\bottomrule
\end{tabular}
\caption{Supervised transformer layers with lowest alignment and highest shift.}
\label{tab:transformer_dissimilar}
\end{table}

\begin{table}[!htbp]
\centering
\small
\begin{tabular}{llll}
\toprule
Criterion & Model.Layer & Value \\
\midrule
Lowest CKA-Linear & dinov2\_small.block1 & 0.0983 \\
Lowest CKA-Linear & dinov3\_vit7b16\_pretrain.block1 & 0.1757\\
Lowest CKA-Linear & dinov2\_base.block1 & 0.2032\\
Lowest RSA Spearman & dinov2\_small.block1 & 0.1498\\
Lowest RSA Spearman & dinov3\_vitb16\_pretrain.block1 & 0.2293\\
Lowest RSA Spearman & dinov2\_base.block1 & 0.3557\\
Largest W1 & dinov3\_vit7b16\_pretrain.block39 & 704.448\\
Next largest W1 & dinov3\_vit7b16\_pretrain.block38 & 578.675\\
Next largest W1 & dinov3\_vit7b16\_pretrain.block37 & 572.599\\
\bottomrule
\end{tabular}
\caption{Self-supervised transformer layers with lowest alignment and highest shift.}
\label{tab:dino_dissimilar}
\end{table}

Across families, early convolutional layers (e.g., `conv1`, `conv0`, `stage1`) show the lowest RSA/CKA, while later transformer blocks (ViT-L/16 blocks 19-23) combine high CKA with sizable distributional shifts by Wasserstein. DINOv3 ViT-7B/16 pretrain exhibits extreme late-block shifts, despite nontrivial geometric alignment; an informative dissociation to consider when selecting layers for analysis or adaptation.

\end{document}